# Results of Evolution Supervised by Genetic Algorithms

**Lorentz JÄNTSCHI**[1], **Sorana D. BOLBOACĂ**[2], **Mugur C. Bălan**[1], **Radu E. SESTRAŞ**[3]

[1] Technical University of Cluj-Napoca, Department of Chemistry, 103-105 Muncii Bvd., 400641 Cluj-Napoca, Romania; lori@academicdirect.org; mbalan@temo.utcluj.ro
[2] "Iuliu Haţieganu" University of Medicine and Pharmacy Cluj-Napoca, Department of Medical Informatics and Biostatistics, 6 Louis Pasteur, 400349 Cluj-Napoca, Romania; sbolboaca@umfcluj.ro
[3] University of Agricultural Sciences and Veterinary Medicine Cluj-Napoca, 3-5 Mănăştur, 400372 Cluj-Napoca, Romania; rsestras@usamvcluj.ro

**Abstract**: A series of results of evolution supervised by genetic algorithms with interest to agricultural and horticultural fields are reviewed. New obtained original results from the use of genetic algorithms on structure-activity relationships are reported.

**Keywords**: Genetic Algorithm (GA), Evolution; Genetic operators.

## INTRODUCTION

Simulation of evolution (through different parameters characterizing the sample under development) is a problem insufficiently explored in the literature; genetic algorithms are just one example.

Studies on other key operators for evolution are found in the literature and focus on algorithmic efficiency (seen in terms of speed with which they achieve maximum proximity and global optimum). A collection of representative works of this type is (Martin & Spears, 2001). Thus, various crossover operators are the subject of study in (Prügel-Bennett, 2001), mutation and crossing in (Spears, 2001), and other dynamic parameters in (Droste et al., 2001).

Studies are too often focused on solving difficult problems using genetic algorithms, sometimes dealing with efficiency (execution time, memory resources needed), rarely to the influence of the development of various strategies and objective (and here again especially on algorithm efficiency) and almost never on other parameters characterizing the sample under development.

For linking simulation → optimization a systematic literature search produced only one reference to a monograph (Stender et al., 1994), and literature is much richer but again on the reverse path from simulation to optimization.

## LITERATURE REVIEW: THEORY

A number of doctoral theses have been conducted on the subject of genetic algorithms in all fields of research and concerns on both basic and applied aspects.

A number of doctoral research of fundamental nature have their starting point the thesis (de Jong and Holland, 1975) supported under the guidance of one of the fathers of modern genetic algorithms - John Henry Holland (born February 2, 1929). Holland is an American scientist, Professor of Psychology, Professor of Electrical Engineering and Computer Science at the University of Michigan, Ann Arbor, he is a pioneer in nonlinear science and complex systems.

Based on the optimization problems, the work (de Jong and Holland, 1975) examines the efficiency of genetic algorithm for some classical problems (Figure 1), which is known from the literature that classical optimization algorithms often failed.

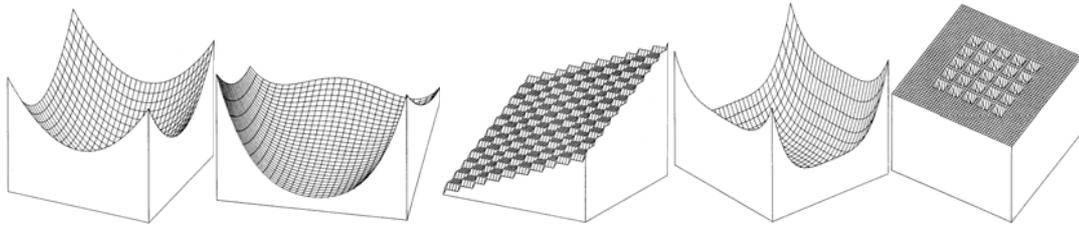

Figure 1. Representation of 3D test functions from F1 to F5 used for assessing of the genetic algorithm in (de Jong and Holland, 1975)'s work

The doctoral research series which (de Jong and Holland, 1975) generated includes study of classical genetic algorithms (GAs) - and here the role of mutation and recombination is the subject of research (Spears and de Jong, 1998) - basically one of very few similar with the one reported here study (addressing the role of selection and survival), a GA modified form - a cooperative co-evolution (Potter and de Jong, 1997; Wiegand and de Jong, 2003), and where the initial population was divided into sub-populations called islands and evolution occurred on each island, allowing however the migration of individuals from one island to another (Skolicki and de Jong, 2007).

A valuable result of (Spears and de Jong, 1998), capitalized sometime later (Spears, 2001) is illustrated in Figure 2 referring the effect of mutation rate on progress towards equilibrium.

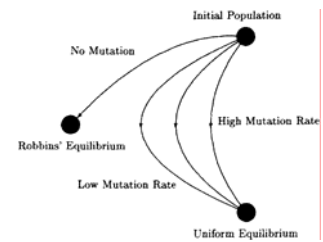

Figure 2. The effect of the mutation on evolution to the equilibrium in (Spears, 2001)'s work

In Figure 2, the Robbins equilibrium (Robbins, 1918) concerns a population that is recombined with no selection and mutation and the balance evenly cover a population that is repeatedly moved without selection and crossing that leads to a point where the objective function value is canceled (Solan, 2009).

Other doctoral research focused on basic research components include various adjustments made on genetic algorithms to solve special classes of difficult problems holding attention: multi-objective optimization (Zitzler and Thiele, 1999), the Pareto optimization (Knowles and Corne, 2002) and multi-agent systems (Panait and Luke, 2006).

Genetic algorithms have exceeded the boundaries of informatics domain due to the potential recovery of the computer simulation results.

Thesis with the objective of designing genetic algorithms, evolutionary programming, and implementation of studies based on them are found in practically all fields of research. Further representative works are detailed in this respect.

## LITERATURE REVIEW: APPLICATIONS

In the field of agriculture, the GA have found their usefulness in crop planning (Matthews and Kraw, 2001), construction on soil erosion risk assessment (Osman and McManus,

2007), in bioengineering to effectively control pollution in the catchments (Veith and Wolfe, 2002), in chemistry in the design of controlled sensory (Dai and Lodder, 2007).

In economics GA were able to solve optimization problems with multiple options (Aickelin and Dowsland, 1999), to manage the multi-scale modeling processes (Sastry et al., 2007), to do mechanical optimization of composite structures (Gantovnik and Gürdal, 2005), and to provide solutions to environmental problems for water quality control strategy (Tufail and Ormsbee, 2006).

Finally, but not least, in biology, two lines come off in terms of development and use of genetic algorithms: the problems of development (Suzuki and Iwasa, 1998) and phylogenetic studies (Zwickl and Hills, 2006).

On the purely applicative, the use of genetic algorithms in agriculture and horticulture, genetic algorithm were found applications in plant growing studies (Venard and Vaillancourt, 2006), on taxonomic classification (Sarmiento-Monroy and Sharkey, 2006) and analysis of genetic diversity (Zhang and Ghabrial, 2006).

THE USE OF GENETIC ALGORITHMS ON STRUCTURE-ACTIVITY RELATIONSHIPS

Optimization problem chosen for the study, namely the structure-activity relationships are at the junction of chemistry with computer sciences and biology. Continuous development of knowledge deposits like those provided by the NIH (National Institute of Health, USA), such as PubMed, PubChem, Genome, etc. stresses the need to have effective tools to articulate this deposited knowledge, and the structure-activity relationships are one of these instruments.

A genetic algorithm (GA) had been developed and implemented in order to identify the optimal solution in term of determination coefficient and estimation power of a multiple linear regression approach for structure-activity relationships. The Molecular Descriptors Family for structure characterization of a sample of 206 polychlorinated biphenyls with measured octanol-water partition coefficients was used as case study.

Probability Distribution Functions (PDFs) and Cumulative Density Functions (CDFs) for a series of observables recorded during GA supervised evolution to the global optimum were seeking in a experimental design in which 46 independent executions were taken into account on every selection and survival strategy as is depicted in Table 1 below.

Table 1.
Simulation results

| Selection* | Survival* | Configuration** | Evolution** |
|---|---|---|---|
| Proportional | Proportional | PCB_4044_cfg.txt | PCB_4044_evo.txt |
| Proportional | Deterministic | PCB_2441_cfg.txt | PCB_2441_evo.txt |
| Proportional | Tournament | PCB_9878_cfg.txt | PCB_9878_cfg.txt |
| Deterministic | Proportional | PCB_5108_cfg.txt | PCB_5108_evo.txt |
| Deterministic | Deterministic | PCB_6369_cfg.txt | PCB_6369_evo.txt |
| Deterministic | Tournament | PCB_6690_cfg.txt | PCB_6690_evo.txt |
| Tournament | Proportional | PCB_5828_cfg.txt | PCB_5828_evo.txt |
| Tournament | Deterministic | PCB_4872_cfg.txt | PCB_4872_evo.txt |
| Tournament | Tournament | PCB_1758_cfg.txt | PCB_1758_evo.txt |

*There are following pairs of selection and survival strategies (PP, PD, PT, DP, DD, DT, TP, TD, TT)
**Files available at: http://l.academicdirect.org/Horticulture/GAs/MLR_MDF_selection_vs_survival

During the research conducted in (Jäntschi and Sestraş, 2010), following conclusions were drawn:
÷ The use of molecular descriptors families on multiple linear regression opens a natural pathway to do the optimization of the regression by using of a genetic algorithm;

- The classical type of genetic algorithm designed and implemented evolutes relatively fast near to the optimum (in the conducted experiment PDF & CDF of the determination coefficient were obtained; probabilities from CDF to obtain 99% from the optimum in 1000 generations are: TD - 55%, PD - 67%, PP - 68%, TP - 73%, PT - 78%, TT - 80%, DD - 87%, DP - 95%, DT - 97%);
- Evolution using different selection and survival strategies create populations of genotypes living in the evolution space with different diversity and variability; under a series of criteria of comparisons (number of genotypes, number of phenotypes, number of associations in regressions, top of 23 occurrences from 46 runs of above listed, etc.), these populations were proof to be grouped and the groups were showed to be statistically different one to each other;
- The investigated evolution objective (determination coefficient of the multiple regressions to maximum) was found to be distributed by the Fisher-Tippett law of extreme values;
- Obtaining of the distribution laws given the opportunity to construct the Lucky lottery and the Unlucky lottery relative to the chosen strategy of selection and survival;
- The relative moments of evolution were found to be distributed by a one parameter degeneration of log-Pearson of type III curve, and two pairs of relatives (for relative moments of evolution) were found in strategies (PP & TT and TD & PD);
- Number of evolutions were found to be distributed by a Fisher-Tippett (again) distribution;
- The dominance in the Fisher-Tippett distributions of evolution objective are Weibull type III extreme values excepting DP strategy which have dominance of Fréchet type II extreme values during evolution;
- The Fisher-Tippett distributions of number of evolutions are Weibull type III extreme values (again) excepting TP strategy, which have a Fréchet type II extreme values distribution.
- The used number of evolutions the variance between strategies were found significantly smaller ($4.07^2$) than the variance inside strategies ($9.68^2$).